\documentstyle[a4,colacl,parsetree,graphics,epsfig]{article}
\begin{document}

\title{Estimation of Stochastic
Attribute-Value Grammars using an Informative Sample}
\author{Miles Osborne \\ osborne@let.rug.nl \\ Rijksuniversiteit Groningen, The Netherlands\thanks{Current
    address: osborne@cogsci.ed.ac.uk, University of Edinburgh, Division of Informatics, 2
    Buccleuch Place, EH8 9LW, Scotland.
 }}
\maketitle
\newtheorem{sent}{}
\newcommand{\labelRule}[2]{\begin{center}\begin{tabular}{ll} #1 & (#2) \end{tabular} \end{center}}
\newcommand{\lexical}[2]{\begin{tabular}{l} #1 $\mapsto$ #2 \end{tabular}}
\newcommand{\numberSentence}[1]{\begin{sent} \small #1 \end{sent}}
\newcommand{\cat}[1]{$\mbox{#1}$}
\newcommand{\lcat}[1]{$\backslash \mbox{#1}$}
\newcommand{\rcat}[1]{$/ \mbox{#1}$}
\newcommand{\psr}[2]{$#1 \rightarrow #2$}
%
%
\newcommand{\sentence}[1]
     {\begin{flushleft}{\it #1}\end{flushleft}}

\begin{abstract}
We argue that some of the computational complexity associated with
estimation of stochastic attribute-value grammars 
can be reduced by training upon an informative subset of the
full training set.  Results using the parsed Wall Street Journal corpus 
show that in some 
circumstances, it is possible to obtain better estimation results 
using an informative sample than when training upon all the available material. Further experimentation demonstrates that with unlexicalised models, a 
Gaussian prior can reduce overfitting.  However, when models are
lexicalised and contain overlapping features,
overfitting does not seem to be a problem, and a Gaussian prior makes
minimal difference to  performance. Our approach is applicable for situations
when there are an infeasibly large number of parses in the training
set, or else for when recovery of these parses from a
packed representation is itself computationally expensive.
\end{abstract}

\section{Introduction}
Abney showed that
attribute-value grammars cannot be modelled adequately  using statistical
techniques which assume that  statistical dependencies are accidental
\cite{Abn97}.  Instead of using a model class that assumed
independence, Abney
suggested  using {\it Random Fields Models} (RFMs) for
attribute-value grammars.  RFMs deal with the graphical  structure of
a parse.  Because they do not make independence assumptions 
 about the stochastic generation
process that might have produced some parse,  they are 
able to  model correctly dependencies that exist within
parses.

When
estimating standardly-formulated RFMs, it is necessary to sum over
all  parses licensed by
the grammar.  For many broad coverage natural language grammars, this
might involve summing over an
exponential number of parses.  This would  make the task computationally
intractable. Abney, following the lead of Lafferty {\it et al},
 suggested a Monte Carlo simulation as a way of reducing the
computational burden associated with RFM estimation \cite{Laff97}.
However, 
Johnson
{\it et al}  considered the form of sampling used in this simulation
(Metropolis-Hastings)  intractable \cite{John99}.  Instead, they 
proposed an alternative  strategy that redefined the
estimation task.  It was argued that this redefinition
made estimation  computationally simple enough that a Monte Carlo
simulation  was
unnecessary.   They presented results obtained using a 
small unlexicalised model trained on a modest corpus. 

Unfortunately, Johnson {\it et al} assumed it was
possible to retrieve {\it all} parses licensed by a grammar when
parsing a given training set.   For us, this
was not the case. In our
experiments with a manually written broad coverage Definite Clause
Grammar (DCG) \cite{Bris96}, we were  only able to 
 recover all parses for
Wall Street Journal sentences that were at most $13$ tokens long
within acceptable time and space bounds on
 computation.   When we used
an incremental Minimum Description Length (MDL) based learner
to extend the coverage of our manually written grammar 
(from roughly $60\%$ to around $90\%$ of the parsed Wall Street Journal), 
the situation became worse.  
Sentence ambiguity considerably
increased.  We were then only able to  recover all parses for
Wall Street Journal sentences that were at most $6$ tokens long \cite{Osbo99a}.

We {\it can} however, and usually in polynomial time, 
 recover  up to $30$ parses for sentences up to $30$ tokens
long when we use a probabilistic unpacking mechanism
 \cite{Carro92}. (Longer sentences than $30$ tokens can be parsed, but the
 number of parses we can recover for them drops off
 rapidly).\footnote{We made an attempt to determine the maximum number
 of parses our grammar might assign to sentences.  On a $450$MHz Ultra Sparc 80
 with $2$ Gb of real memory, with a limit of  at most  $1000$
 parses per sentence, and allowing no more than $100$ CPU seconds per
 sentence, we found that sentence ambiguity increased exponentially
 with respect to sentence length.  Sentences with $30$ tokens had an estimated
 average of $866$ parses (standard deviation $290.4$).  Without the
 limit of $1000$ parses per sentence, it seems likely that this
 average would increase.}
 However, $30$ is far less than the maximum number of
 parses per sentence our grammar might assign to  Wall Street Journal 
sentences.  Any training set we have access to will therefore be
 necessarily limited in size.

We therefore need an estimation strategy that takes seriously the
issue of extracting the best performance 
from a limited size training set. A limited size training set
means one created by retrieving at most $n$ parses per
sentence. Although we cannot recover all possible parses, we do have a
choice as to which parses estimation should be based upon.

Our approach to the problem of making RFM estimation 
feasible for our highly ambiguous DCG is to seek out an {\it
informative}
 sample
and train upon that. We do not redefine the estimation task in a
non-standard way, nor do we use a Monte Carlo simulation.

We call a sample {\it informative} if it both leads to the selection of a
model that does not underfit or overfit, and also is typical of future
samples. Despite one's intuitions, an informative sample might be a
proper subset of the full training set.  This means that estimation
using 
the informative sample might yield better results than  estimation
using all of the training set.

The rest of this paper is as follows. Firstly we  introduce RFMs.
Then we show how they may be estimated and how an informative sample
might be identified. Next, we give details of the
attribute-value 
grammar we use, and show how we go about modelling it.  We then
present two sets of experiments.  The first set is small scale, and
are designed to show the existence of an informative sample. The
second set of experiments are larger in scale, and  build
upon the computational savings we are able to achieve using a
probabilistic unpacking strategy.  They show how large models (two
orders of magnitude larger than those reported by Johnson {\it et al})
can be estimated using the parsed Wall Street Journal
corpus. Overfitting is shown
to take place.  They also show how
this overfitting can be (partially) reduced by using a Gaussian prior.
Finally, we end with some
comments on our work.

\section{Random Field Models}
Here we show how attribute-value grammars may be modelled using
RFMs. Although our commentary is in terms of RFMs and grammars,
it should be obvious that RFM technology can be applied to other
estimation scenarios.

Let $G$ be an attribute-value grammar, $D$ the set of sentences  within the
string-set defined by $L(G)$ and $\Omega$ the union of the set of
parses assigned
to each sentence in $D$ by the grammar $G$.
A Random Field Model, $M$, consist of two components: a set of {\it
features}, $F$ 
  and a set of {\it weights}, $\Lambda$. 

Features
   are 
the basic building blocks of RFMs.  They enable the system designer to
   specify the key aspects of what it takes to differentiate one parse
   from another parse.
Each feature
is a 
function from a
parse to an integer. 
Here, the integer
value associated with a
feature is interpreted as the number of times a feature
   `matches' (is `active') 
with a parse.  
 Note features should not be confused
with features as found in feature-value bundles (these will be called
attributes instead). Features are usually manually
  selected by the system designer. 

The other component of a RFM,  $\Lambda$, is 
 a set of weights.  
Informally, weights tell us how features are to be used  when
 modelling parses.  For example, 
an active  feature with a large weight  might
indicate that some  parse had a high probability.  Each weight
 $\lambda_i$ is associated with a  feature $f_i$.  
Weights are real-valued numbers  and are  
 automatically determined by an estimation process (for example using Improved
Iterative Scaling \cite{Laff97}).  One of the nice properties of RFMs
 is that the likelihood function of a RFM is strictly concave.  This means that
there are no local minima, and so we can be be sure that scaling will
result in estimation of  a RFM that is globally optimal.

The (unnormalised) total weight 
of a parse $x$, $\psi(x)$, is a function of the $k$ features  
that are `active' on a 
parse:
\begin{eqnarray}
\psi(x) = \exp(\sum_{i=1}^{k}\lambda_{i}f_{i}(x))
\end{eqnarray}

The probability of a parse, $P(x \mid M)$, is simply the result of
normalising the total weight associated with that parse:
\begin{eqnarray}
P(x \mid M) = \frac{1}{Z}\psi(x)  \label{additive} \\ 
Z = \sum_{y \in \Omega} \psi(y)
\end{eqnarray}
The interpretation of this probability depends upon the
application of the RFM.  Here, we use parse probabilities to
reflect preferences for parses.

When using RFMs for parse selection, we simply select the parse that
maximises $\psi(x)$.  In these circumstances, there is no need to
normalise (compute $Z$). Also, when computing $\psi(x)$ for competing
parses, there is no built-in bias towards shorter (or longer)
derivations, and so no need to normalise with respect to derivation
length.\footnote{The reason there is no need to normalise with respect
to derivation length is that features can have positive or negative
weights. The weight of a parse will therefore not always monotonically
increase with respect to the number of active features.}

\section{RFM Estimation and Selection of the Informative Sample \label{iis}}
We now sketch how RFMs may be estimated and then outline how we seek
out an informative sample.  

We use Improved Iterative Scaling (IIS) 
 to estimate RFMs. In
outline, the IIS algorithm is as follows:
\begin{enumerate}
\item Start with  a reference distribution $R$, a set of features $F$ and a
set of weights $\Lambda$. Let $M$ be the RFM defined using $F$ and $\Lambda$.
\item Initialise all weights to zero. This makes the initial model uniform.
\item Compute  the expectation of each feature w.r.t $R$.
\item For each feature $f_i$ \label{here}
\begin{enumerate}
\item Find a weight $\check{\lambda_i}$ that equates the expectation of $f_i$ 
w.r.t $R$ and the expectation of $f_i$ w.r.t $M$.
\label{exp}
\item Replace the old value of $\lambda_i$ with $\check{\lambda_i}$.
\end{enumerate}
\item If the model has converged to $R$, output $M$.
\item Otherwise, go to step \ref{here}
\end{enumerate}
The key step here is \ref{exp}, computing the expectations of
features w.r.t the RFM.  This involves calculating the probability of
a parse, which, as we saw from equation \ref{additive}, requires a
summation over all parses in $\Omega$.

We seek out an informative sample $\Omega_{t}$ ($\Omega_{t} \subseteq \Omega$) as follows:
\begin{enumerate}
\item Pick out from $\Omega$ a sample of size $n$. \label{there}
\item Estimate a model using that sample and evaluate it.
\item If the model just estimated 
shows signs of overfitting (with respect to an
unseen held-out data set), halt and output the model. 
\item Otherwise, increase $n$ and go back to step \ref{there}.
\end{enumerate}

Our approach is motivated by the following (partially related) observations:
\begin{itemize}
\item Because we use a non-parametric model class and select an
instance of it in terms of some sample (section
\ref{modelling} gives details), a stochastic complexity argument tells
us that an overly simple model (resulting from a small sample) is
likely to underfit.  Likewise, an overly complex model (resulting from
a large sample) is likely to overfit.  An informative sample will
therefore relate to a model that does not under or overfit.
\item On average, an informative sample will be `typical' of future
samples. For many real-life situations, this set is likely to be small
relative to the size of the full training set. 
\end{itemize}
We incorporate the first observation through our search
mechanism. Because we start with small samples and gradually increase
their size, we remain within the domain of efficiently recoverable
samples. 

The second observation is (largely) incorporated in the way we pick
samples. The experimental section of this paper goes into the relevant
details.

Note our approach is heuristic: we cannot afford to evaluate all
$2^{\mid \Omega \mid}$ possible training sets.
The actual size of the informative sample $\Omega_{t}$ will depend
both the upon the model class used and the maximum sentence length we
can deal with.  We would expect richer, lexicalised models
to exhibit overfitting with smaller samples than would be the case
with unlexicalised models.  We would expect the size of an informative
sample to increase as the maximum sentence length increased.

There are similarities between our approach and with estimation using
MDL \cite{Riss89}.
However, our implementation
does not explicitly attempt to minimise code lengths.  Also, there are
similarities with importance sampling approaches to RFM 
estimation (such as  \cite{Chen99b}).  However,
such attempts do not
minimise under or overfitting.

\section{The Grammar}
The grammar we model with Random Fields,
(called the {\it Tag Sequence Grammar} \cite{Bris96}, or TSG for
short) was
developed with regard to coverage, and 
when compiled consists of 455 Definite Clause Grammar (DCG) rules.  It does 
not parse sequences of
words
directly, but instead assigns derivations to sequences of part-of-speech 
tags (using the CLAWS2 tagset.  
  The grammar is relatively shallow, (for
example, it does not fully analyse unbounded dependencies)  but it does
make an attempt to deal with  common constructions, such as dates or names,
commonly found in corpora, but of little theoretical interest.  Furthermore,
it integrates into the syntax a text grammar, grouping utterances into
units that reduce the overall ambiguity.  

\section{Modelling the Grammar \label{modelling}}
Modelling the TSG with respect to the parsed Wall Street Journal
consists of two steps: creation of a feature set and definition of
the reference distribution.

Our feature set is created by parsing sentences in 
the training set ($\Omega_{T}$),
and using each parse to
instantiate {\it templates}. Each template defines a family of
features.  At present, the templates we
use are somewhat ad-hoc.  However, they are motivated by the
observations that linguistically-stipulated units (DCG rules) are
informative, and that many DCG applications in preferred parses can be
predicted using lexical information. 

The first template creates
features that count the number of times a DCG
instantiationis present
within a parse.\footnote{Note, all our features suppress any terminals
that appear in a local tree. Lexical information is included when we
decide to lexicalise features.}  For example, suppose we  parsed  the Wall Street Journal AP:
\numberSentence{unimpeded by  traffic}

\begin{figure}
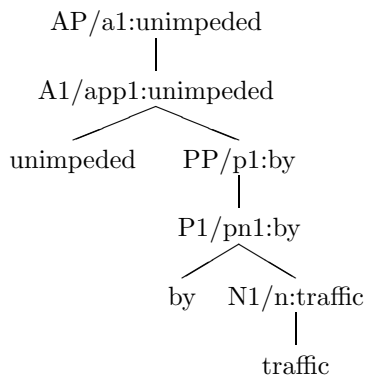

\begin{small}
\begin{center}
\begin{parsetree}     
(.AP/a1:unimpeded.
  (.A1/app1:unimpeded. .unimpeded. (.PP/p1:by. (.P1/pn1:by. .by. (.N1/n:traffic.
 .traffic.)))))
\end{parsetree}
\end{center}
\caption{TSG Parse Fragment \label{tsg1}}
\end{small}
\end{figure}

A parse tree generated by TSG might be as shown in figure \ref{tsg1}.
Here, to save on space, we have labelled each interior node in the
parse tree with TSG rule names, and not attribute-value bundles.
Furthermore,  we have annotated each node with the head word of the phrase in
question. Within our grammar, heads are (usually) explicitly
marked. This means we do not have to make any guesses when identifying the head
of a local tree. With head information, we are able to lexicalise
models.  We have suppressed tagging information. 

For example, a  feature  defined using this template 
might count the number of times the we saw: 
\begin{center}
\begin{parsetree}(.AP/a1. .A1/app1.) \end{parsetree}
\end{center}
in a parse.
Such features record some of the context of the rule application, in
that rule applications that differ in terms of how attributes are bound
will be modelled by different features.

Our second template creates features that are partially lexicalised.
For each local tree (of depth one) that has a PP daughter, we create a
feature that counts the number of times that local tree, decorated
with the head-word of the PP, was seen in a parse.   An example of
such a lexicalised feature would be:
\begin{center}
\begin{parsetree}(.A1/app1. .PP/p1:by.) \end{parsetree}
\end{center}
These features are designed to model PP attachments that can be
resolved using the head of the PP.

The third and final template creates features that are again partially
lexicalised.  This time, we create local trees of depth one that are
decorated with the head word.  For example, here is one such feature:
\begin{center}
\begin{parsetree}(.AP/a1:unimpeded. .A1/app1.) \end{parsetree}
\end{center}

Note  the second and third templates result in features that
overlap with features resulting from applications of the first template.

We create the reference distribution $R$ (an association of probabilities with
TSG parses of sentences, such that the probabilities reflect parse preferences)
using the following process:
\begin{enumerate}
\item Extract some sample $\Omega_{T}$ (using the approach mentioned in section \ref{iis}).
\item For each sentence in the sample, for
each parse of that sentence, compute the `distance' between the TSG parse
and the WSJ reference parse.  In our approach, distance is calculated 
 in terms of
a weighted sum of crossing rates, recall and precision.  Minimising
it maximises our definition of parse plausibility.\footnote{Our
distance metric is the same one used by Hektoen \cite{Hek97b}}
However, there is nothing inherently crucial about  this decision.  Any
other objective
function (that can be represented as an exponential distribution) could
be used instead.
\item Normalise the distances, such that for some sentence, the sum of
the distances of all recovered TSG parses  for that sentence is a constant
across all sentences.  Normalising in this manner ensures that each
sentence is  equiprobable (remember that RFM probabilities are in terms
of parse preferences, and not probability of occurrence in some corpus).
\item Map the normalised distances into probabilities.  If $d(p)$ is
the normalised distance of TSG parse $p$, then associate with parse
$p$ the reference probability given by the maximum likelihood
estimator: 
\begin{eqnarray}
\frac{d(p)}{\sum_{x \in
\Omega_t}d(x)}
\end{eqnarray}
\end{enumerate}
Our approach therefore gives partial credit (a non-zero reference
probability) to all parses in $\Omega_{t}$.  $R$ is therefore not as
discontinuous as  the equivalent distribution used by Johnson {\it
et al}. We therefore do not need to use simulated annealing or other
numerically intensive techniques to estimate models.
\section{Experiments}
Here we present two sets of experiments.  The first set demonstrate
the existence of an informative sample.  It also shows some of the
characteristics of three sampling strategies.  The second set of
experiments is larger in scale, and show RFMs (both lexicalised and
unlexicalised) estimated using sentences up to $30$ tokens long.
Also, the effects of a Gaussian prior are demonstrated as a way of
(partially) dealing with overfitting.

\subsection{Testing the Various Sampling Strategies \label{small}}
In order to see how various sizes of sample  related to
estimation accuracy and whether we could achieve similar levels of
performance without recovering all possible parses, we ran the
following experiments.

We used a model consisting of features that were defined using all
three templates.  We also threw away all features that occurred less
than two times in the training set.    We randomly split the Wall Street
Journal into  disjoint training, held-out and testing sets.  All sentences in
the training and held-out sets were at most $14$ tokens
long. Sentences in the testing set were at most $30$ tokens long.  There
were $6626$ sentences in the training set,  $98$ sentences in the
held-out set and $441$ sentences in the testing set.  Sentences in the
held-out
 set had on average $12.6$ parses, whilst sentences in the testing-set
 had  on average $60.6$ parses per sentence.

The held-out set was used to decide which model performed best.
Actual performance of the models should be judged with  respect to the testing set.

Evaluation was in terms of exact match: for each sentence in the test
set, we awarded ourselves a point if the RFM ranked highest the same
parse that was ranked highest using the reference probabilities.
 When
evaluating with respect to the held-out set, we 
recovered all parses for sentences in the held-out set. When
evaluating with respect to the testing-set, we recovered at most
$100$ parses per sentence.

For each run, we ran IIS for the
same number of iterations (20). In each
case, we evaluated the RFM after each other iteration and recorded the
best classification performance. This step was designed to avoid
overfitting distorting our results.

Figure \ref{results2} shows the results we obtained with possible ways
of picking `typical' samples.  The first column shows the maximum number of parses per
sentences that we retrieved in each sample.  

The second column shows the size of the sample (in parses).

The other columns give classification accuracy results (a percentage)
with respect to the testing set.  In parentheses, we give performance
with respect to the held-out set.

The column marked {\it Rand} shows the performance of runs that 
used a sample
that contained parses which were  randomly and uniformly selected out of the
set of all possible parses.   The
classification accuracy results for this sampler    are averaged over 10
runs. 

The column marked {\it SCFG} shows the results obtained when using a
sample that   contained  parses that were  retrieved using the
probabilistic unpacking strategy.  This did not involve retrieving all
possible parses for each sentence in the training set.  Since there is
no random component, the results are from a single run. Here, parses
were ranked using a stochastic context free backbone approximation of
TSG.  Parameters were estimated using simple counting.

Finally, the column marked {\it Ref} shows the results obtained when
using a sample that contained the overall $n$-best parses per
sentence, as defined in terms of the reference distribution.  

\begin{figure}
\begin{center}
\begin{small}
\begin{tabular}{ll|lll}
Max parses & Size & Rand  & SCFG & Ref\\ \hline
  1                 &  6626            &     25.2 (51.7) & 23.3 (50.0)
  & 23.4 (50.0)  \\      
           2                 &  12331            &  37.9 (63.0) & 40.4
  (60.3) &  40.4 (60.0) \\
           3                 &  17026            &  43.2 (65.5)    &
  43.7 (63.8) &  43.7 (63.8)\\
           5                 &  24878            &  43.7 (70.2)      &
  45.8 (69.5) &  45.8 (69.5)\\
           {\bf 10}                &  {\bf 39581}            &  {\bf 47.4 (72.0)}   &
  {\bf 47.0 (70.0)} &  {\bf 46.9 (70.0)}\\
           100                & 119694             &  45.0 (68.7)
  &  45.0 (68.0) &  45.0 (68.0)\\ 
           1000                & 246686             &  44.4 (67.4)
  &  43.0 (67.0) &  43.0 (67.0)\\ 
           $\infty$                &  267400            &  43.0 (66.0)
  & 43.0 (66.0)  & 43.0 (66.0) 
\end{tabular}
\end{small}
\end{center}
\caption{Results with various sampling strategies \label{results2}}
\end{figure}

As a baseline, a model containing randomly assigned weights produced a 
classification accuracy of $45\%$ on the held-out sentences. These results
were averaged over $10$ runs.

As can be seen, increasing the sample size produces better results
(for each sampling strategy).  Around a  sample size of  $40k$ parses,
overfitting starts to manifest, and performance bottoms-out.  One of
these is 
therefore our informative sample. Note
that the best  sample ($40k$ parses) is less than $20\%$ of the total
possible training set.

The difference between the various samplers is marginal, with a slight
preference for {\it Rand}. However the fact that {\it SCFG} sampling
seems to do almost as well as {\it Rand} sampling, and furthermore does not require unpacking
all parses, makes it the sampling strategy of choice.  

{\it SCFG} sampling is biased in the sense that the sample produced
using it
will tend to concentrate around those parses that are all close to the
best parses.  {\it Rand} sampling is unbiased, and, apart from the
practical problems of having to recover all parses, might in some
circumstances be better than {\it SCFG} sampling.  
At the time of
writing this paper, it was unclear whether we could combine {\it SCFG}
with {\it Rand} sampling -sample parses from the full distribution
without unpacking all parses.  We suspect that for probabilistic
unpacking to be efficient, it must rely upon some non-uniform
distribution. Unpacking randomly and uniformly would probably
result in a large loss in computational efficiency.

\subsection{Larger Scale Evaluation}
Here we show results using a larger sample and testing set. We also 
show the effects of lexicalisation, overfitting,
and overfitting avoidance using a Gaussian prior.  Strictly speaking
this section could have been omitted from the paper.  However, if one
views estimation using an informative sample as overfitting avoidance,
then estimation using a Gaussian prior can be seen as another,
complementary  take on
the problem.

The experimental setup was as follows.  We randomly split the Wall Street
Journal corpus into a training set and a testing set. Both sets contained
sentences that were at most $30$ tokens long.  When creating the set
of parses used to estimate RFMs, we used the SCFG approach, and
retained the top 25 parses per sentence. Within the training set (arising
from $16,200$ sentences),
there were $405,020$ parses.  The testing set consisted of $466$
sentences, with an average of $60.6$ parses per sentence.

When evaluating, we retrieved at most
 $100$ parses per sentence in the testing set and 
 scored them using our reference
distribution. As before, we awarded ourselves a point if the most
probable testing parse (in terms of the RMF) coincided with the most
probable parse (in terms of the reference distribution).  In all
cases, we ran IIS for $100$ iterations.

For the first experiment, we used just the first template (features
that related to DCG instantiations) to create model 1; the second
experiment used the first and second templates (additional features
relating to PP attachment) to create model 2.  
The final
experiment used all three templates 
(additional features that were head-lexicalised) to create model 3.

\begin{flushleft}
\begin{figure}[!hbt]
\psfig{figure=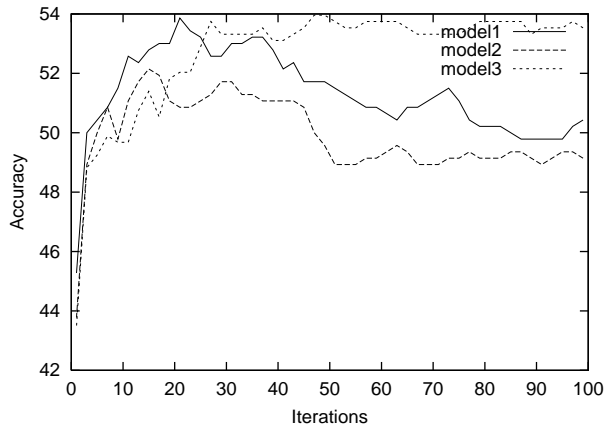}
\caption{Classification Accuracy for Three Models Estimated using
Basic IIS \label{big1}}
\end{figure}
\end{flushleft}

The three models contained $39,230$, $65,568$ and $278,127$ features respectively,

As a baseline, a model containing randomly assigned weights achieved a
$22\%$ classification accuracy. These results were averaged over $10$
runs. Figure \ref{big1} shows the classification accuracy using models 1, 2
and 3.
\begin{center}
\begin{figure}[!hbt]
\psfig{figure=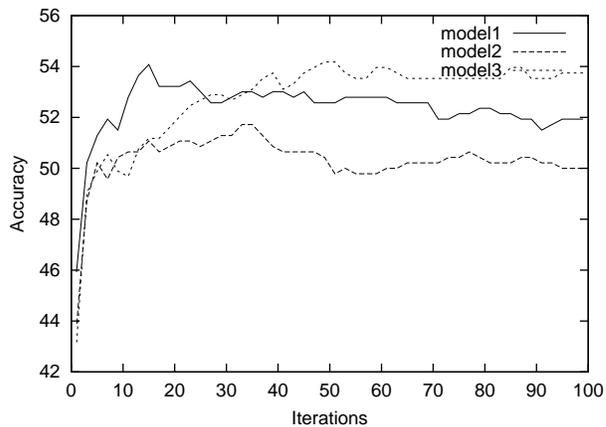}
\caption{Classification Accuracy for Three Models Estimated using a
Gaussian Prior and IIS \label{big2}}
\end{figure}
\end{center}

As can be seen, the larger scale experimental results were better than
those achieved using the  smaller samples (mentioned in section
\ref{small}).  The reason for this was because we used longer
sentences.  The informative sample derivable from such a training set
was likely to be larger (more representative of the population) than
the informative sample derivabled from a
training set using shorter, less syntactically complex sentences. 
With the unlexicalised model, we see clear signs of overfitting.
Model 2 overfits even more so.  For reasons that are unclear, we see
that the larger model 3 does not appear to exhibit overfitting.

We next used the Gaussian Prior method of Chen and Rosenfeld to reduce
overfitting \cite{Chen99}. This involved 
integrating a Gaussian prior (with a zero mean) into IIS and searching
for the model that maximised the product of the likelihood and prior probabilities.
For the experiments reported here, we used a single variance over the
entire model (better results might be achievable if multiple variances
were used, perhaps with one variance per template type).
The actual value of the variance was found by
trial-and-error.  However, optimisation using a held-out set is easy
to achieve.

We repeated the large-scale experiment, but this time using a Gaussian
prior. Figure \ref{big2} shows the classification accuracy of the 
models when using a Gaussian Prior.  

When we used a Gaussian prior, we found that all models showed signs
of improvement (allbeit with varying degrees): performance either
increased, or else did not decrease with respect to the number of
iterations. Still, model 2 continued to underperform.  Model 3 seemed
most resistent to the prior.  It therefore appears that a Gaussian prior is
most useful for unlexicalised models, and that for models built from
complex, overlapping features, other forms of smoothing must be used instead.

\section{Comments}
We argued that RFM estimation for broad-coverage attribute-valued
grammars could be made computationally tractable by training upon an
informative sample. Our small-scale experiments suggested that using
 those parses that could be efficiently unpacked ({\it SCFG}
sampling) was almost as effective as sampling from all possible
parses ({\it Rand} sampling). Also, we saw that models should not be
both built and also estimated using all possible parses. Better results can be
obtained when models are built and trained using an informative sample.

Given the relationship between sample size and model complexity, we
 see that when there is a danger of overfitting, one should build models
on the basis of an informative  set.  However, this leaves open the
 possibility of training such a model upon a superset of the informative set.
Although we have not tested this scenario, we believe that this would
 lead to better results than those achieved here.

The larger scale experiments showed that RFMs can be estimated using
relatively long sentences. They also showed that a simple Gaussian
prior could reduce the effects of overfitting.  However, they also
showed that excessive overfitting probably required an alternative
smoothing approach.

The smaller and larger experiments can be both viewed as
(complementary) ways of
dealing with overfitting. We conjecture that of the two approaches, 
the informative sample
approach is preferable as it deals with overfitting directly:
overfitting results from fitting to complex a model with too little
data. 

Our ongoing research will concentrate upon stronger ways of dealing
with overfitting in lexicalised RFMs. One line we are pursuing is to
combine a compression-based prior with an exponential model.  This
blends MDL with Maximum Entropy. 

We are also looking at alternative template sets.  For example, we
would probably benefit from using templates that capture more of the
syntactic context of a rule instantiation.
\section*{Acknowledgments}
We would like to thank Rob Malouf, Donnla Nic Gearailt and the
anonymous reviewers for comments.
This work was supported by 
the TMR Project {\it Learning Computational Grammars}.  

\bibliographystyle{/users1/osborne/latex/acl}

\end{document}